\pdfoutput=1
\documentclass[11pt]{article}
\usepackage[margin=1.1in]{geometry}
\usepackage{booktabs}

\usepackage{caption}
\usepackage{pgfplots}
\pgfplotsset{compat=1.18}
\usepackage[hidelinks]{hyperref}
\usepackage{parskip}

\title{Evaluation Choices Decide the Forecasting Leaderboard:\\
Evidence from a Production Marketplace Panel}

\author{Md Rezwanul Islam \quad Wael Mohammed\\[0.5em]
Field Nation LLC, Minneapolis, MN, USA\thanks{Corresponding author:
\texttt{rezwanul.islam@fieldnation.com} (co-author:
\texttt{wael.mohammed@fieldnation.com}). ORCID: Md Rezwanul Islam
\texttt{https://orcid.org/0009-0002-4100-3796}; Wael Mohammed
\texttt{https://orcid.org/0009-0005-4211-8976}.}}
\date{August 2026}

\begin{document}
\maketitle

\begin{abstract}
A forecasting benchmark reports which method won. We show that the answer is set by the
evaluator's choices before any model is fitted. We benchmark 24 forecasting methods and one
textbook reference, including six 2025-era time series foundation models, on a production
marketplace panel of 1,887 business customers over 67 months. We hold the data, the horizon and the period fixed, and vary only
the evaluation design. Three choices each reverse or dissolve a headline conclusion.
Changing the unit of analysis from the market total to the individual customer moves our
production baseline from second of nineteen, beaten by nothing, to twenty-third of
twenty-five. Nineteen of its twenty-four challengers beat it there. Changing how much error is pooled decides whether a
Diebold--Mariano test finds anything at all. Scoring prediction intervals rather than point
forecasts reorders the field almost completely, with a rank correlation of 0.02 on intermittent
demand. We then measure what the deployed system gets from this. Its selection rule
captures 55\% of the distance between doing nothing and choosing with hindsight. The reversal is
not a quirk of our data. We ran the released protocol, unchanged, on the public M5 retail panel.
The same baseline shape places first at the market total and last per series,
beaten by everything, and a replayed selection rule closes 64.7\% of the same floor-to-ceiling
distance there. Adding five zero-shot foundation models to that roster changes who wins at the
total, not the shape. The
bands' blind spot travels too: conformal bands under-cover most on the
spikiest items. Splitting our own panel into ever smaller groups turns the contrast
into a curve: the baseline's rank worsens at every level of disaggregation. We release the
evaluation protocol and report an error of our own that inverted a result before we caught it.
\end{abstract}

\section{Why this paper exists}
\label{sec:intro}

Forecasting benchmarks are published to answer a question: which method should I use? The reader
is meant to look at the winner and act on it. This paper argues that the winner is often a
property of the benchmark rather than of the methods.

We are not the first to say something like this. There is a careful literature on how forecast
evaluation goes wrong, covering error measures, data partitioning and statistical
testing \cite{hewamalage2023pitfalls}. There is also work measuring how unstable a competition's
own ranking is, which finds aggregation to be one of the main
causes \cite{hewamalage2021m5setup}. Our claim is narrower and it is measured. We
take one production panel and one fixed set of methods. We change nothing about the data, the
horizon, or the period. We change only decisions the evaluator makes, and we record how far the
conclusion moves. It moves a long way. We then ask whether that is a fact about our firm or a
fact about the design, by running the same released protocol on a public retail panel
(Section~\ref{sec:m5}). It is the design.

This is a companion to earlier work in a different field \cite{islam2026qml}. That paper
benchmarked quantum and classical machine learning for power-system attack detection and reached
the same conclusion there: eight evaluation choices each reversed or moved a result with the
models held fixed. The present paper tests the same idea on forecasting, which is a different problem with
different conventions. Two domains agreeing is the reason to take the idea seriously.

The paper's method is the protocol itself, and we release it as a runnable artifact. It holds
a panel fixed, changes one evaluation choice at a time, and measures how far the reported
winner moves. The same artifact replays a deployed selection rule against a floor and a
ceiling. The results sections of this paper are its outputs on two panels.

We also report something more useful to a practitioner than a leaderboard. A deployed system
cannot wait for the field to agree. It has to pick a method every month and serve a number. We
measure what our rule for doing that is actually worth, against a floor and a ceiling.

\section{The setting and the data}
\label{sec:setting}

Our firm runs a two-sided marketplace for on-site technical field service. Customers, whom we
call buyers, post work orders. Independent technicians accept and complete them. The number the
business plans around is gross transaction value, or GTV. It is the total value of work completed
on the platform each month.

Two facts about this data shape everything below. First, there is no subscription floor. Every
dollar is earned again each month from project work. Second, and more important here, the data is
two forecasting problems sharing one name. The market total is smooth, because thousands of
buyers rise and fall independently and largely cancel out. The individual buyer series are
spiky, often short, and dominated by a few large accounts.

The panel is monthly GTV by buyer. It runs from January 2021 to July 2026 and contains 5,779
buyers. Most of them are not forecastable, and we consider that the correct answer rather than a
gap. Our system needs 13 months of trading history before it will forecast a buyer at all,
because a year-over-year method has no base below that. We would rather return nothing than
invent a number. Roughly 2,200 buyers clear that floor, and 1,887 of them clear it at one or more
of our evaluation origins. The rest form an abstention cohort: the system serves nothing
for them, and this benchmark scores nothing for them. A leaderboard that quietly dropped them
would describe a smaller and easier business than the real one, so we state the counts instead.

Buyers differ enormously in how they buy, so we classify them. We use the standard demand
taxonomy, which splits series on two axes: how often demand occurs, and how variable the amount
is when it does \cite{syntetos2005croston}. The standard cut-offs apply: an average demand
interval of 1.32 months, and a squared coefficient of variation of 0.49. That gives four
classes: Smooth, Erratic, Intermittent and Lumpy. We recompute a buyer's class at every evaluation origin using only that
origin's training window. Classifying on the full series would let the future leak into the
grouping.

\section{What we compare, and how}
\label{sec:design}

We score 24 forecasting methods plus one textbook reference. They cover the families a
practitioner would consider today. Six are classical statistical methods. Four are
intermittent-demand methods built for spiky series. Five are machine-learning models trained
across the whole buyer panel at once, three gradient-boosted trees and two neural networks. Six
are time series foundation models, used with no training on our data at all: three from the
Chronos family \cite{ansari2024chronos}, TimesFM~2.5 \cite{das2024timesfm}, Moirai~2.0
\cite{woo2024moirai}, and FlowState. Three are simple baselines.

The last method is the one that runs in production. It is last year's same month, adjusted for
recent momentum. We claim no credit for it. It is the seasonal naive with drift that enterprise
systems have shipped for decades, listed in Oracle's JD~Edwards documentation as forecasting
method~2 \cite{oracle2024jde}. We also score a plain seasonal naive with no drift, as the
reference the scale-free metrics divide by.

Every method implementation is the production system's own. We did not re-implement anything for
this study. That removes the most common source of benchmark error, which is a re-coded method
that differs subtly from the deployed one.

Evaluation uses rolling origins, the standard design for testing forecasts out of
sample \cite{tashman2000}. How to evaluate across time without letting the future leak into the
past is its own literature, and shuffled cross-validation is where it goes wrong
\cite{bergmeir2012}. We take eight cutoffs at three-month spacing, each forecasting
the following six months, each scored only against actuals that did not exist at the cutoff. Five
of the methods are trained across the panel rather than per series, and those are retrained at
every cutoff on data strictly before it. Training them once on the full panel would leak every
test window into their training set.

The six-month horizon deserves a note. The business consumes a twelve-month forecast. We benchmark
at six months because that is the block length the deployed selection rule itself cross-validates
on, so this is the horizon at which methods are actually chosen.

We checked whether the horizon matters by rerunning the whole comparison at twelve months. Only
cutoffs with a full twelve months of realized data afterwards qualify, so the grid shrinks to five
origins and 7,200 blocks. One model, Chronos T5-Large, is excluded from this check on compute
grounds: it takes roughly fifty times longer per series than any other foundation model here. At
six months it ranked twentieth of twenty-five, far from the comparisons the check is testing.
The ranking is essentially unchanged: the rank correlation with the six-month ranking is 0.98, no
method moves more than three places, and the same six methods occupy the top of both tables. The
production baseline places twenty-third of twenty-four rather than twenty-second, with sixteen
methods beating it significantly against eighteen at six months on the same roster. Errors are uniformly higher at twelve months, as
expected, but the ordering is not what changes.

We report two error measures: MASE, the mean absolute scaled error, and RMSSE, the root mean
squared scaled error \cite{hyndman2006measures}. Each divides a method's error by the plain
seasonal naive's error on the same basis, so a score below one beats that reference. Reporting
two is deliberate. They punish errors in different
ways, so agreement between them is evidence and disagreement is a warning. That the choice of
error measure decides which forecast is best is itself established: each measure rewards a
forecast aimed at a different target \cite{kolassa2020}. Significance against the
production baseline uses the Diebold--Mariano test on per-step absolute errors, with the
small-sample correction \cite{diebold1995,harvey1997}. Comparisons across all methods at once use the Friedman
test with the Nemenyi critical difference \cite{demsar2006}. Because we run 24 tests inside each
demand class, we control the false discovery rate \cite{benjamini1995fdr}.

Support is not uniform, and we report it rather than hide it. Some methods need more history than
some buyer--origin pairs carry. Four methods therefore score 10,341 of the 11,516 blocks and the
rest score all of them. The Friedman test runs on complete blocks only, and we state how many were
dropped.

One assumption in that test needs care. Friedman treats its blocks as independent, and ours are
not: a single buyer contributes up to eight buyer--origin blocks, which are correlated. The
critical difference we report is therefore narrower than it should be. We checked how much this
matters by rebuilding the test on one randomly chosen origin per buyer, which makes the blocks
independent by construction. The critical difference widens from 0.37 to 0.92, as expected on a
sixth of the blocks. The ranking itself barely moves: across twenty random draws the rank order
agrees with the full ranking at a mean Spearman correlation of 0.98. We report the conservative
critical difference alongside the ranking for this reason.

\section{Result one: the unit of analysis reverses the conclusion}
\label{sec:inversion}

\begin{table}[t]
\centering
\caption{The same panel, the same window, scored two ways. Changing the unit of analysis moves
the production baseline from second place to near-last and turns "no method beats it" into "most
methods beat it"; the flip survives on the shared roster and origins (Section~\ref{sec:inversion}).
Significance is Diebold--Mariano against the production baseline, Harvey--Leybourne--Newbold
corrected; direction is the sign of the DM statistic on the pooled per-step loss.}
\label{tab:inversion}
\begin{tabular}{lrr}
\hline
 & Aggregate & Per buyer \\
\hline
Series scored & 1 & 1,887 \\
Rolling origins & 22 & 8 \\
Forecasts compared & 19 methods & 25 methods \\
Blocks & 22 & 11,516 \\
Ranked on & mean WAPE & mean RMSSE \\
\hline
Rank of the production baseline & 2 of 19 & 23 of 25 \\
Methods significantly better & 0 & 19 \\
Methods significantly worse & 10 & 4 \\
Per-step losses pooled by DM & 126 & 66,068 \\
\hline
\end{tabular}
\end{table}

Exhibit~\ref{tab:inversion} is the paper in one table. The left column scores the market total.
The right column scores individual buyers. The data, the horizon and the period are the same, and
the unit of analysis is the headline difference. Three smaller differences ride along, because
each column is the evaluation the production system actually runs at that level. The aggregate
column scores nineteen methods at twenty-two monthly origins. Five of the six methods it lacks are
trained across many series and do not exist for a single total; the sixth is a simple per-buyer
baseline. The per-buyer column scores all twenty-five at eight quarterly origins. And the two
columns rank on different error measures: weighted absolute percentage error, or WAPE, for the single
total, and RMSSE across the buyers. None of the three is doing the work. We recomputed both
columns on exactly the shared roster and the shared origins: nineteen methods, eight cutoffs,
nothing else different. The baseline again places second of nineteen on the total, and
seventeenth of nineteen per buyer. Re-ranking the total on RMSSE moves the baseline up, from
second to first. The inversion is the unit of analysis, not the roster, not the origin grid, and
not the error measure.

On the market total the production baseline places second of nineteen, and no method beats it by
a statistically distinguishable margin. On individual buyers the same baseline places
twenty-third of twenty-five, and nineteen of twenty-four methods beat it. A practitioner reading
only the first result would conclude that method choice does not matter and stop looking. A
practitioner reading only the second would conclude that it matters enormously. Both would be
reading a real result from the same firm in the same quarter.

The mechanism is not mysterious, and it is not new. Aggregation cancels the swings of individual
buyers, and what survives is a smooth seasonal series that almost anything can fit. The panel puts a
number on that cancellation: in a typical month the market total moves 7\% of its average
level, and the
median scored buyer moves 32\%. Disaggregation exposes
exactly the variation that separates methods. That the level matters is the founding observation of
the hierarchical forecasting literature \cite{athanasopoulos2024hierarchical}, and that
competition rankings shift under different tests has been shown before
\cite{koning2005m3}. Closest to us is a study of the M5 competition's own evaluation setup. It
measures how much a ranking moves across equivalent datasets at fixed models, and finds
aggregation to be a main driver of that movement \cite{hewamalage2021m5setup}. That result is
about how stable a ranking is within a level. Ours is about the ranking reversing between two
levels. The two fit together, and theirs implies that the aggregate half of our own comparison is the
less reliable half, which supports our reading.

Our contribution is not the mechanism. It is the size of the effect, measured
on one production panel with everything else held fixed: the choice is rarely stated, and here it
decides the reported winner entirely.

One more objection deserves its own check. Scoring the total does not only change the unit. It
also weights every buyer by its dollar size, where the per-buyer table counts every buyer
equally, so the inversion could in principle be a re-weighting effect in disguise. It is not.
Re-ranking the per-buyer table with each buyer weighted by its average monthly volume moves the
baseline from twenty-third to twentieth of twenty-five, with nineteen methods still ranked
ahead of it. Pooling absolute errors over dollars rather than buyers gives the same picture.
Disaggregation does the work, not equal weighting.

\begin{table}[t]
\centering\footnotesize
\setlength{\tabcolsep}{4pt}
\caption{Per-buyer accuracy over 8 rolling origins and
11,516 buyer--origin blocks. Ranked on mean RMSSE, the
metric the deployed system selects on. "Beats" counts blocks in which the method's RMSSE is
below the production baseline's. $p$ is Diebold--Mariano against that baseline, two-sided,
Benjamini--Hochberg adjusted within this family of 24
tests. For LightGBM (global) and Seasonal trend the significant difference favors the baseline: a worse pooled
per-step loss despite the lower mean RMSSE.
Nemenyi critical difference 0.37 on
10,341 complete blocks.
$^{\dagger}$ scored on 10,341 blocks rather than
11,516: these methods require more history than some buyer--origin pairs carry.}
\label{tab:per_buyer}
\begin{tabular}{rlrrrrrr}
\hline
 & Method & \multicolumn{2}{c}{RMSSE} & MASE & Mean & Beats & $p$ \\
 & & mean & median & mean & rank & & (BH) \\
\hline
1 & DeepAR (global)$^{\dagger}$ & 0.806 & 0.523 & 0.932 & 10.44 & 7,753 & $<$0.0001 \\
2 & N-HiTS (global)$^{\dagger}$ & 0.809 & 0.536 & 0.929 & 10.64 & 7,662 & $<$0.0001 \\
3 & TimesFM 2.5 & 0.831 & 0.519 & 0.926 & 10.43 & 8,693 & $<$0.0001 \\
4 & Moirai 2.0 & 0.833 & 0.539 & 0.929 & 10.65 & 8,592 & $<$0.0001 \\
5 & IMAPA & 0.838 & 0.557 & 0.975 & 10.33 & 8,666 & $<$0.0001 \\
6 & FlowState & 0.842 & 0.553 & 0.950 & 10.98 & 8,760 & $<$0.0001 \\
7 & EWMA (3-month) & 0.850 & 0.571 & 0.988 & 10.76 & 8,412 & $<$0.0001 \\
8 & Chronos-2 & 0.852 & 0.549 & 0.947 & 11.35 & 8,216 & $<$0.0001 \\
9 & TSB & 0.865 & 0.566 & 1.009 & 11.66 & 8,007 & $<$0.0001 \\
10 & AutoTheta & 0.868 & 0.594 & 1.015 & 11.76 & 8,050 & $<$0.0001 \\
11 & Median-12 & 0.868 & 0.563 & 0.979 & 12.20 & 8,038 & $<$0.0001 \\
12 & Croston SBA & 0.875 & 0.560 & 1.026 & 12.08 & 7,836 & $<$0.0001 \\
13 & Croston & 0.876 & 0.573 & 1.031 & 12.27 & 7,789 & $<$0.0001 \\
14 & AutoETS & 0.879 & 0.586 & 1.035 & 11.65 & 8,036 & $<$0.0001 \\
15 & Chronos-Bolt & 0.880 & 0.562 & 0.980 & 12.33 & 7,774 & $<$0.0001 \\
16 & CatBoost (global) & 0.901 & 0.577 & 0.996 & 13.59 & 7,467 & 0.0096 \\
17 & XGBoost (global) & 0.902 & 0.580 & 0.996 & 14.16 & 7,345 & $<$0.0001 \\
18 & LightGBM (global) & 0.908 & 0.574 & 1.002 & 13.39 & 7,443 & 0.0096 \\
19 & Linear trend & 0.908 & 0.662 & 1.063 & 13.39 & 7,291 & 0.1546 \\
20 & Chronos T5-Large & 0.962 & 0.611 & 1.078 & 15.76 & 6,456 & $<$0.0001 \\
21 & MSTL & 0.964 & 0.691 & 1.116 & 15.11 & 6,688 & $<$0.0001 \\
22 & Seasonal trend$^{\dagger}$ & 1.062 & 0.832 & 1.229 & 17.51 & 4,722 & $<$0.0001 \\
23 & \textbf{Seasonal naive + momentum (production)} & 1.090 & 0.794 & 1.216 & 16.84 & 0 & -- \\
24 & Seasonal naive (plain, no drift) & 1.100 & 0.830 & 1.235 & 17.71 & 4,968 & $<$0.0001 \\
25 & Prophet$^{\dagger}$ & 1.231 & 0.885 & 1.426 & 18.02 & 4,446 & $<$0.0001 \\
\hline
\end{tabular}
\end{table}

Exhibit~\ref{tab:per_buyer} gives the full per-buyer ranking. The mean and median columns tell
the same story, so the ordering is not the work of a few extreme blocks. One reading note on the
significance column: twenty-two methods sit above the baseline, and nineteen of them beat it on
the pooled test. Two of the other three, LightGBM and Seasonal trend, are significant in the
opposite direction --- a better mean rank, a worse pooled per-step error. Which way a significant
difference points depends on the loss it is measured on, which is this paper's point in
miniature. A buyer-level bootstrap
says it is not the work of a few large accounts either: in 1,992 of 2,000 redraws of the panel
the baseline places exactly twenty-third, and it never moves more than one place. First place
stays inside the two global neural models in every draw. Two observations are
worth pulling out. Foundation models and global neural models occupy the top of the table, which is consistent
with the case for learning across many related series \cite{monteromanso2021global}. And Prophet finishes last of twenty-five.

\subsection{A second choice hiding inside the first}

There is a further choice buried in the significance testing, and it deserves separating.

The Diebold--Mariano test pools per-step forecast errors. At the market total we pool 126 of
them, and the test finds no method significantly better than the baseline. Per buyer we pool tens of thousands, and
the test finds almost everything significant at $p < 0.0001$. It is the same test on the same
methods. How much error the evaluator chooses to pool decides what it reports.

Pooled errors are also not independent. Buyers share calendar months, and one demand shock
touches thousands of blocks at once, while the small-sample correction we use handles dependence
over time, not across buyers. As a check, we averaged the error differences across buyers
within each calendar month and re-ran the same test on the twenty-five monthly averages. All
nineteen methods that beat the baseline stay significant. On our margins the dependence changes nothing.
On a panel with thinner margins it could, and the check is cheap.

This is a known property of any test whose power grows with sample size, but it has a practical
consequence for benchmark readers. At a large enough panel, statistical significance stops
carrying information, because everything is significant. That is why we report Nemenyi mean-rank
separation alongside the tests. A critical difference tells you whether methods are far apart
relative to the spread, which does not inflate with the number of series.

\subsection{From a contrast to a curve}
\label{sec:dose}

\begin{table}[t]
\centering\small
\caption{From a contrast to a curve. Every buyer in the panel is assigned to one of $k$
size-sorted groups, using only data from before the first origin, and the shared roster of
nineteen methods is scored on each group's aggregate over the same origins. The baseline's
rank worsens monotonically as the aggregate is split: Spearman against $\log k$ is
1.00 on Friedman mean rank and
0.95 on the other two conventions. The last column
shows why: splitting the total replaces smooth aggregates with spiky ones, so the
aggregation level and the demand mix are one dial seen from two sides.}
\label{tab:dose}
\begin{tabular}{rrrrrrr}
\hline
Groups & Buyers & Blocks & \multicolumn{3}{c}{Baseline rank of 19, by convention} & Smooth \\
$k$ & per group & & median WAPE & mean RMSSE & Friedman & share \\
\hline
1 & 5,779 & 23 & 2 & 5 & 5 & 100\% \\
5 & 1,156 & 103 & 14 & 14 & 12 & 57\% \\
25 & 231 & 479 & 14 & 14 & 14 & 51\% \\
125 & 46 & 2,336 & 16 & 17 & 16 & 50\% \\
625 & 9 & 9,862 & 16 & 17 & 17 & 27\% \\
\hline
\end{tabular}
\end{table}

The comparison so far has two points: the market total and the individual buyer. Two points
leave a question open. Is there a cliff somewhere between them, or does the conclusion decay
smoothly as the aggregate is split? Exhibit~\ref{tab:dose} answers with a dial. We assign every
buyer in the panel to one of $k$ groups by size, using only data from before the first origin.
The shared roster of nineteen methods is then scored on each group's aggregate, for $k$ from 1
to 625. At $k=1$ the group is the market total. At $k=625$ a group holds about nine buyers.

The baseline's rank never improves at any step of the dial, on any of the three ranking
conventions. On
Friedman mean rank the ordering is perfectly monotone: fifth at the total, twelfth at five
groups, fourteenth at twenty-five, sixteenth at one hundred twenty-five, seventeenth of
nineteen at six hundred twenty-five. Most of the fall has already happened by $k=5$, where a
group still sums more than a thousand buyers. The cancellation that protects the baseline at
the total is fragile: even a mild split destroys most of it.

The last column of the exhibit says why, and connects this result to the next section. At
$k=1$ the aggregate is Smooth. By $k=625$ only 27\% of the group aggregates are, and the rest
are Erratic, Lumpy or Intermittent. Splitting the panel does not merely shrink each series. It
changes what kind of series the evaluator is scoring. The unit of analysis and the demand mix
are one dial seen from two sides.

That forecast accuracy depends on the aggregation level is an established theme in
supply-chain forecasting, and the field's advice is to choose the level, or combine levels,
rather than assume one \cite{babai2022aggregation}. The curve reads the same dial differently.
With the roster and everything else held fixed, it is the ranking of methods, not only their
accuracy, that moves with the level, and it moves monotonically.

The grouping rule is not doing the work either. Re-run with buyers assigned to groups at
random instead of by size, the curve stays perfectly monotone on the WAPE convention: ranks
2, 6, 10, 15, 16 across the same five levels. Random groups each hold a share of the large
buyers, so they are smoother, and the baseline's fall arrives later than under size sorting.
Composition moves where the fall happens. Splitting the aggregate is what causes it.

\begin{figure}[t]
\centering
\begin{tikzpicture}
\begin{axis}[
    width=0.88\textwidth, height=0.5\textwidth,
    xmode=log, log ticks with fixed point,
    xlabel={Number of aggregates the panel is split into (log scale)},
    ylabel={Baseline's place in its roster},
    ytick={0,0.5,1}, yticklabels={first,middle,last},
    ymin=-0.06, ymax=1.06, y dir=reverse,
    legend pos=north east, legend cell align=left, legend style={font=\small},
    grid=major, major grid style={dotted},
]
\addplot+[mark=*, thick, color=black] coordinates {(1,0.222) (5,0.611) (25,0.722) (125,0.833) (625,0.889)};
\addlegendentry{Production panel: 19 methods, size-sorted groups}
\addplot+[mark=square*, thick, dashed, color=black!55] coordinates {(1,0.000) (3,0.000) (10,0.000) (30,0.500) (70,0.500) (30490,1.000)};
\addlegendentry{M5: 7 methods, its own hierarchy}
\end{axis}
\end{tikzpicture}
\caption{The same fall on two panels. The production baseline's place in its roster slides
from near the top toward the bottom as the aggregate is split, whether the groups are
constructed by size on our panel or are M5's own levels, from the market total down to the
single item--store series. Rank is shown as position within each roster, first to last,
because the rosters differ in size. Production points are Friedman mean-rank placements at
$k$ = 1 to 625 groups (Exhibit~\ref{tab:dose}); M5 points are mean-RMSSE placements at
its six native levels.}
\label{fig:curves}
\end{figure}
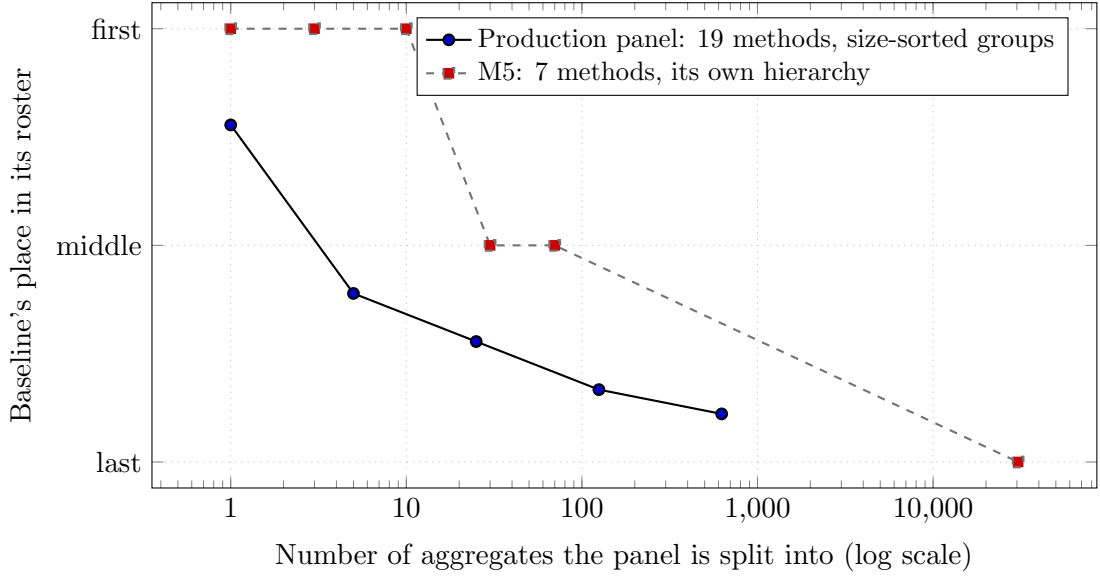

Nor is the curve a property of constructed groups. The public M5 panel of
Section~\ref{sec:m5} carries a real hierarchy, and running the same protocol down it gives
the same dial with nothing constructed. The baseline ranks first at the market total, and
first across the three state totals and the ten store totals. It ranks fourth across thirty
category--store and seventy department--store aggregates, and last across the 30,490
item--store series. The rank correlation with the level's size is 0.93, and 0.94 on the
Friedman convention; ranking the levels on median error agrees at the bottom and wiggles
near the top. The pooling result rides along: no
challenger separates significantly at any aggregate level, and all six do at the bottom.
A real hierarchy walks the same curve. Figure~\ref{fig:curves} draws both falls side by
side.

The released package closes the loop with a control we can turn all the way. Its synthetic
generator has a spikiness dial, and sweeping it holds everything fixed except how volatile
the entities are. Turn it up from smooth and the entity half of the inversion assembles
first. The baseline slides from first to last per entity, while staying first or second at
the total in forty-four of forty-five panels. Turn it well past the marketplace setting and
the effect starts to dissolve from the other side. The heavy-tailed entity swings stop
canceling, the total itself turns spiky, and single challengers begin to separate at the
aggregate as well. The inversion needs both halves --- spiky entities and a smooth
total --- and the dial shows each half arriving and leaving on its own schedule.

\section{Result two: the winner depends on the demand pattern}
\label{sec:classes}

\begin{table}[t]
\centering\footnotesize
\setlength{\tabcolsep}{4pt}
\caption{Which method wins depends on the buyer's demand pattern. Buyers are classified at every
origin from that origin's training window alone, on average demand interval and squared
coefficient of variation. Three different methods lead the four classes; the pooled leaderboard's
single winner leads only two of them.}
\label{tab:by_class}
\begin{tabular}{lrrlrr}
\hline
Demand class & Blocks & Best & Top three & Baseline & Baseline \\
 & & RMSSE & & rank & RMSSE \\
\hline
Smooth & 4,074 & 0.803 & N-HiTS, DeepAR, Moirai & 22 of 25 & 1.054 \\
Erratic & 4,295 & 0.700 & DeepAR, N-HiTS, TimesFM & 23 of 25 & 1.029 \\
Intermittent & 579 & 1.199 & IMAPA, AutoTheta, TSB & 24 of 25 & 1.447 \\
Lumpy & 2,568 & 0.870 & DeepAR, TimesFM, IMAPA & 24 of 25 & 1.168 \\
\hline
\end{tabular}
\end{table}

Exhibit~\ref{tab:by_class} splits the same comparison by demand class. Pooling the four groups
into a single leaderboard hides real structure: three different methods lead the four classes,
and the pooled winner leads only two of them.

We claim nothing new here. That the best method depends on the features of the series is an
established result, and tailoring the method to the data is the advice that follows from
it \cite{petropoulos2014horses}. What this section adds is the size of the split on one
production panel. One detail is worth noting: the intermittent-demand methods win on
intermittent demand, which is the taxonomy doing what it was built to do.

This matters because pooling is the default. A benchmark that reports one table is implicitly
claiming that its winner is the winner everywhere in the panel. On this data that claim is false.
The practical form of the finding is that a deployed system should route a series to a method
class based on how the series behaves, rather than adopting one champion model.
Section~\ref{sec:selection} measures what that routing is worth against exactly that
alternative.

\section{Result three: scoring intervals reorders the field}
\label{sec:intervals}

\begin{table}[t]
\centering\footnotesize
\setlength{\tabcolsep}{4pt}
\caption{Prediction-interval quality over 193,067 scored records, one
record per method--buyer--origin triple. The coverage columns score the deployed system's own
conformal bands, calibrated at each origin on that buyer's earlier origins only; the last three
columns span all twenty-five methods. $\rho$ is the Spearman correlation between ranking methods
on point accuracy and ranking them on interval quality (mean scaled interval score); "move" is the
largest number of places any method
shifts between the two rankings. The bands are calibrated where demand is regular and
under-cover badly where it is not.}
\label{tab:intervals}
\begin{tabular}{lrrrrrrrr}
\hline
 & & \multicolumn{2}{c}{80\% band} & \multicolumn{2}{c}{95\% band} & \multicolumn{3}{c}{Point vs interval} \\
 & Records & cover & error & cover & error & $\rho$ & $p$ & move \\
\hline
Overall & 193,067 & 0.782 & -1.8 & 0.864 & -8.6 & 0.39 & 0.0509 & 16 \\
Smooth & 71,337 & 0.815 & 1.5 & 0.925 & -2.6 & 0.61 & 0.0011 & 11 \\
Erratic & 73,358 & 0.796 & -0.4 & 0.884 & -6.6 & 0.55 & 0.0047 & 12 \\
Intermittent & 7,848 & 0.714 & -8.6 & 0.741 & -20.9 & 0.02 & 0.9099 & 16 \\
Lumpy & 40,524 & 0.710 & -9.0 & 0.742 & -20.8 & 0.31 & 0.1285 & 16 \\
\hline
\end{tabular}
\end{table}

So far everything has scored point forecasts. Most production systems ship intervals as well, and
ours does. Interval quality is scored by the mean scaled interval score, the interval metric of
the M4 competition \cite{makridakis2020m4}. It charges for the width of the band and adds a
penalty every time the actual lands outside it. Every method's band is built the same way, by
split conformal prediction on that method's own point errors. The comparison therefore isolates
the point forecaster behind the band, not any model's built-in uncertainty output.
Exhibit~\ref{tab:intervals} asks whether the
method you would pick on point accuracy is the method whose intervals you would want.

Largely, it is not. Overall the rank correlation between the two criteria is 0.39, and it is not
significant at the conventional level. On Intermittent demand it is 0.02, which is no relationship
at all. Both are correlations across twenty-five methods, so neither is a precise estimate. The
contrast between the regular classes and the spiky ones is the finding, not the decimals. The
best method on point accuracy ranks eleventh on interval quality. Methods move up to
sixteen places of twenty-five between the two rankings. Where demand is regular the two criteria
agree far better, which is itself informative: the choice matters most exactly where forecasting
is hardest.

The same exhibit contains a finding about our own system that we did not go looking for. Our
conformal prediction bands are well calibrated on Smooth demand, where a nominal 95\% band
delivers 92\%. On Lumpy and Intermittent buyers the same nominal 95\% band delivers about 74\%.
The bands are least trustworthy on precisely the customers whose forecasts are least trustworthy.
We report this because a reader running a similar system should check the same thing. Section~\ref{sec:m5} measures how much of this result
travels beyond our panel.

\section{Result four: what the deployed selection rule is worth}
\label{sec:selection}

\begin{table}[t]
\centering\small
\caption{What the deployed selection rule is worth. At every origin the rule is replayed using
only that buyer's earlier origins, then scored against two references: doing nothing (always the
production baseline) and choosing the best method with hindsight. The last column is the share of
the distance between those two that the rule actually closes. Scored on RMSSE, the metric the rule
ranks on; MASE gives 22.1\% and
48.9\%.}
\label{tab:selection}
\begin{tabular}{lrrrrrrr}
\hline
 & Blocks & Baseline & Served & Oracle & Gain & Wins & Gap \\
 & & RMSSE & RMSSE & RMSSE & & & closed \\
\hline
Overall & 5,832 & 0.946 & 0.724 & 0.540 & 23.5\% & 4,489 & 54.6\% \\
Smooth & 2,313 & 0.995 & 0.783 & 0.572 & 21.3\% & 1,738 & 50.1\% \\
Erratic & 2,207 & 0.883 & 0.667 & 0.489 & 24.5\% & 1,702 & 54.8\% \\
Intermittent & 229 & 1.245 & 0.985 & 0.829 & 20.9\% & 198 & 62.6\% \\
Lumpy & 1,083 & 0.907 & 0.659 & 0.512 & 27.4\% & 851 & 62.8\% \\
\hline
\end{tabular}
\end{table}

A benchmark can afford to say that the answer depends. A production system cannot. Ours has to
choose a method for every buyer every month and serve a number.

The rule it uses is simple enough to state in a sentence. For each buyer we roll through earlier
origins, score every candidate method at each one, and rank them on mean RMSSE. A method must
have scored on at least half the available folds, and its average scaled bias must sit within a
fixed threshold. The three best survivors are averaged, and that average is served.

The rule needs validation history before it can choose, so it is only measurable at origins that
have some. We require at least three earlier origins, which leaves 5,832 of the 11,516
buyer--origin blocks. Exhibit~\ref{tab:selection} measures the rule on those, against two
references. The floor is doing nothing: always serving the production baseline. Practitioners
know this discipline as forecast value added, which scores every step of a forecasting process
against the naive alternative \cite{gilliland2013}; our floor applies it to method selection.
The ceiling is
choosing the single best method at each origin with hindsight, which nobody can do. That ceiling
picks one method at a time. A ceiling that could also average methods, as the rule itself does,
would sit higher still, and every share measured against it would shrink. The rule closes 55\% of the
distance between them, improving on the floor by 23.5\%, and it wins in 77\% of blocks. The pattern holds in every demand
class, and a second metric with a different loss shape gives the same answer.

One more reference belongs beside the floor and the ceiling. The leaderboard invites a simpler
policy than selection: adopt its winner everywhere. Always serving DeepAR, with the baseline
where it lacks the history to forecast, closes 52\% of the same gap; the best method with full
support closes 54\%. The rule's edge over a single champion is therefore real but modest, and
most of the closable gap goes to any strong method served consistently. Per demand class, the
rule wins everywhere except Lumpy, where the champion is itself the leading method.

Whether to pick a method per series or apply one method to all of them has been studied
directly, along with the conditions under which each wins \cite{fildes2015selection}. Why
validation-based selection disappoints is also understood: the method that fits the validation
windows best is often not representative of what comes next \cite{petropoulos2023selection}.
Our question is narrower. We are not asking which selection policy is better in general. We are
asking what one already deployed rule is worth, on its own panel, against a floor and a ceiling.

We think this is the most useful number in the paper. It is not a claim that our rule is optimal.
It is a statement of how much of the available value a reasonable, simple, deployed rule actually
captures, measured rather than asserted.

One detail of the replay deserves a check. Our origins sit three months apart while the horizon is
six months, so consecutive validation folds overlap. The deployed system spaces its own folds six
months apart, and they do not. We therefore repeated the whole replay with the folds thinned to
match production, which makes them disjoint. The result holds and moves slightly in favor of the
rule: it then closes 55\% of the gap and improves on the floor by 26\%, on a smaller set of 2,022
blocks. Overlapping folds were not doing the work.

\section{Does the result travel? The same protocol on public panels}
\label{sec:m5}

\begin{table}[t]
\centering\small
\caption{The reversal is not ours alone: the released protocol, run end-to-end on the public
M5 (Walmart) panel aggregated to monthly, with the protocol's 7 reference methods and the
seasonal-naive-with-momentum baseline. The verdict is computed by the released code, not
asserted. Significance is Diebold--Mariano on the protocol's pooled squared-error
differentials, Benjamini--Hochberg adjusted. Under the non-overlapping-fold variant (origin
spacing equal to the horizon) and re-ranked on MASE instead of RMSSE the placements are
unchanged (166,809 blocks in that variant).}
\label{tab:m5}
\begin{tabular}{lrrr}
\hline
 & Market total & Store totals & Per item--store \\
\hline
Series scored & 1 & 10 & 30,449 \\
Rolling origins & 20 & 20 & 8 \\
Blocks & 20 & 200 & 232,651 \\
\hline
Rank of the baseline & 1 of 7 & 1 of 7 & 7 of 7 \\
Methods significantly better & 0 & 0 & 6 \\
Per-step losses pooled by DM & 120 & 1,200 & 1,395,906 \\
\hline
\end{tabular}
\end{table}

Everything above is one firm's panel, and a fair reader should ask whether the reversal is a
property of our data or of the design choice. The supplementary package gives that question a
mechanical answer, because the protocol it contains runs on any panel. We ran it, unchanged, on
the public M5 retail dataset: the daily unit sales of 30,490 item--store series from ten Walmart
stores \cite{makridakis2022m5}, summed to monthly over the 63 complete months. The panel has the
same two-level structure as ours. Item--store series are spiky and often zero; the market total
is smooth. The protocol's own seven reference forecasters stand in for the roster, and its
baseline carries the same shape as our production baseline: last year's value, scaled by recent
momentum.

Exhibit~\ref{tab:m5} shows the same reversal, sharper. At the market total the baseline places
first of seven, and no method beats it by a distinguishable margin. The ten store totals give
the same answer. Per item--store, across 232,651 blocks, the same baseline places last of
seven, and all six challengers beat it at $p<0.001$. The verdict is printed by the released
code from the run itself, not asserted by us. The finer results travel too. The winner differs
by demand class, and the baseline is last in all four classes. The pooled Diebold--Mariano test
finds nothing at the total and everything per series. And the selection-rule replay works on
data it was not designed for. Replayed per item--store, the rule closes 64.7\% of the distance
between always serving the baseline and choosing with hindsight. It wins in 77\% of 101,729
blocks. On our panel the same rule closes 55\%.

Two robustness checks repeat the checks we ran at home. Spacing the origins a full horizon
apart, so no validation folds overlap, leaves every placement unchanged. Ranking on MASE
instead of RMSSE leaves every placement unchanged as well. The replay survives the same
treatment: with no overlapping folds it still closes 60.5\% of the floor-to-ceiling
distance, on 56,772 blocks.

The M5 competition's own results are the mirror image of this table, and the two agree. Its
top entrants, global models trained across the whole hierarchy, beat the benchmarks by the
most at the top levels and by only a few percent per item \cite{makridakis2022m5}. Which side
wins at a level depends on what the roster contains. Their roster held the strongest global
learners. Ours here holds simple references, and on our own production roster the global
neural models lead per buyer (Section~\ref{sec:inversion}). What both settings share is this
paper's point: the level decides the margins before the methods do.

We then measured that claim rather than leaving it as an argument. Five of the six
foundation models are zero-shot, so nothing but compute excluded them from this panel, and
we added them to the roster. The exception is Chronos T5-Large, roughly fifty times slower
per series, which we excluded on compute grounds as before. The winner at the total changes:
FlowState takes first place and beats the baseline by a distinguishable margin, the only
method that does, and the baseline slips to second of twelve. The shape does not change. Per
item--store the baseline is last of twelve, beaten by eleven, and the extended M5 table now
mirrors the production table almost exactly: second at the total, last per series.
Exhibit~\ref{tab:m5_ext} gives the placements. One
detail is worth a sentence. The best method per item on this panel is not a foundation model
but damped simple smoothing, ahead of TimesFM and FlowState by a nose. One caution attaches
to these five rows. M5 is public, and public panels can sit inside a foundation model's
pretraining data; audits have documented exactly this leakage for several model
corpora \cite{li2026tsfmaudit}. We cannot rule out that a model here saw M5 in training.
Exposure would flatter the foundation-model rows, so the one result it could not manufacture
is the baseline finishing last per item. Our own panel is private, so its rows carry no such
risk. These rows land in a
live argument. Whether large pretrained models beat simple methods at forecasting is
currently contested \cite{tan2024llm}, and shared benchmarks are being built to settle
it \cite{aksu2024gifteval}. Our result says what those benchmarks will find depends on the
unit and the pooling their designs fix first. The design question comes before the model
question.

\begin{table}[t]
\centering\small
\caption{The M5 roster extended with the five zero-shot foundation models (Chronos T5-Large
excluded on compute grounds), scored on the same grid as Exhibit~\ref{tab:m5} with the same
tests. The winner at the total changes --- FlowState, the only method that beats the baseline
by a distinguishable margin there --- and the shape does not: the baseline is second at the
total and last per item--store, the production panel's exact pattern.}
\label{tab:m5_ext}
\begin{tabular}{lrrr}
\hline
 & Market total & Store totals & Per item--store \\
\hline
Blocks & 20 & 200 & 232,651 \\
Rank of the baseline & 2 of 12 & 4 of 12 & 12 of 12 \\
Methods significantly better & 1 & 3 & 11 \\
Winner & FlowState & FlowState & Damped smoothing \\
\hline
\end{tabular}
\end{table}

The interval result travels in halves. The released code builds rolling conformal bands of
the same design on the M5 roster, and the calibration failure follows the data. The baseline's nominal
95\% band delivers 86\% coverage on Smooth items, 79\% on Intermittent and 68\% on Lumpy.
It is the same blind spot as at home, on someone else's panel. Interval forecasting is hard
enough that the M5 competition gave it a separate track \cite{makridakis2022m5uncertainty}.
The reordering half weakens here. Across seven
similar reference methods the point and interval rankings largely agree, at a rank
correlation of 0.89 overall against 0.39 across our twenty-five production methods. The
agreement is weakest on Intermittent items at 0.68 --- the same gradient as at home. Our
first reading was that mixing method families is what breaks the agreement. We tested that
reading, and it failed. Extending this roster to twelve methods across families --- the
five zero-shot foundation models beside the seven references --- moves the overall
correlation from 0.89 to 0.89. A mixed roster is not enough on its own. The reordering
therefore rests on the production setting, whose panel and roster both still differ ---
including models trained on the panel itself --- and we do not claim to have located its
cause. The miscalibration needs no such care: it is a property of spiky data and follows
the data everywhere. These are correlations across seven and twelve methods, so the
contrast with home, not the decimals, is the read.

A third panel marks the mechanism's boundary. Australian tourism is the reconciliation
literature's standard hierarchy \cite{athanasopoulos2009tourism}: 304 bottom series of
monthly visitor nights by region and
purpose, and they are seasonal and smooth, with none of the spikiness of items or buyers.
The mechanism says aggregation has little to cancel when the entities are already smooth,
so it predicts a weaker reversal here. That is what we measure. The baseline is first of
seven at the national total, and it falls to last of seven per series by rank --- but only
one of six challengers separates significantly, against all six on M5. The fade is not a
sample-size artifact. M5 thinned to the same 304 series keeps the baseline last in every
draw, and three to six of the six challengers stay significant. The fade belongs to
the data, not to the count. The effect fades
exactly where the mechanism says it should. A mechanism that predicts where its own effect
fades is easier to trust than one that always fires.

One boundary of these replications is stated plainly: they use the protocol's reference
methods plus the zero-shot foundation models, not the full production roster, so they
replicate the design effect, not our leaderboard. That is the right test. The claim of this
paper is not that any particular method wins. It is that the evaluation design decides what
wins, and that claim now has the same answer on two unrelated panels, with a measured
boundary on a third.

\section{What we would tell a team starting this work}
\label{sec:guidance}

Five things follow from the above, and none of them require agreeing with our leaderboard.

State the unit of analysis before reporting a winner. On this panel it is the single largest
determinant of the answer, larger than any difference between methods.

Report rank separation, not only $p$-values, once the panel is large. At tens of thousands of
pooled errors, significance is nearly automatic and stops discriminating.

Split the leaderboard by demand class before naming a winner. On this panel three different
methods lead the four classes, and the pooled winner leads only two of them.

Score intervals separately if you ship intervals. On our production roster the point-accuracy
ranking does not transfer, and on the hardest series it carries no information about interval
quality at all. On the public panel the two rankings agree far better, so the transfer failure
is a risk to check rather than a law --- and checking it costs one rolling calibration run.

Measure the selection rule, not only the methods. The question a business is actually asking is
what the system is worth, and that is a different measurement from which model has the lowest
error.

\section{Limitations}
\label{sec:limits}

The full benchmark is one firm, one panel, and one industry. The central result no longer rests
on that panel alone. The unit-of-analysis reversal, the class dependence, the pooling effect and
the selection-rule payoff all reproduce on the public M5 panel through the released protocol,
and the tourism panel marks the mechanism's boundary (Section~\ref{sec:m5}). But those
replications use the protocol's reference methods plus five zero-shot foundation models, not
the full production roster, and a handful of panels is still a handful. The twenty-five-method
leaderboard remains single-firm evidence. The interval result is replicated only in half: its
calibration failure reproduces on M5. Its reordering does not appear on the public panel under
either roster we tried --- seven similar references, or twelve methods spanning families ---
and so remains production-only evidence, with its cause not yet located.

We also looked for a panel statistic that would say in advance how strongly the unit of
analysis will move a leaderboard. Two designs failed their pre-registered tests, and the
failed designs ship in the supplementary package beside the ones that worked. As it stands,
the dial of Section~\ref{sec:dose} shows when the effect appears; no single number we found
predicts how hard.

The headline tables are scored at six months, the horizon at which our system selects. The
twelve-month check in Section~\ref{sec:design} reproduces the ranking, but on a shorter grid of
five origins and without one of the twenty-four methods. It is a robustness check, not a second
full benchmark.

Two of the twenty-four methods train neural networks without a fixed random seed. Across two independent full runs
every deterministic method reproduced to three decimal places, and one neural method moved in the
third. Readers should treat those two rows as carrying that much noise.

\section{An error we made}
\label{sec:error}

We report this because a paper about evaluation choices should hold itself to the standard it
recommends.

Our first run of the selection-rule analysis reported the opposite of Section~\ref{sec:selection}.
It found the deployed rule to be substantially worse than doing nothing, with an apparently
overwhelming test statistic. The cause was a scale denominator. RMSSE divides by the seasonal
naive's root mean squared error, and we had scored the rule's blend using the seasonal naive's
mean absolute error instead. The second quantity is always the smaller of the two, so the blend's
score was inflated by a constant factor while the methods it was compared against were not. The
entire effect was that factor.

The lesson generalizes beyond our mistake. A scale-free metric is only comparable when every
quantity being compared was scaled the same way. This is easy to violate the moment a study scores
a combination, such as an ensemble or a rule, alongside single methods. It is also invisible
to a plausibility check, because the wrong answer was not obviously absurd. It is the reason this
paper reports two metrics rather than one.

\section*{Disclosures}

\textbf{Declarations of interest.} The authors are employed by Field Nation LLC, which operates the business-to-business service marketplace whose production system is analyzed in this paper. Field Nation approved the publication of this work. The authors declare no other competing interests.

\textbf{Data-owner consent.} Field Nation LLC consented to the publication of the aggregate statistics (rates, ratios, and counts) derived from its proprietary production panel.

\textbf{Funding.} This research received no external funding.

\textbf{Data and code availability.} The underlying marketplace data is proprietary and cannot be released. No absolute revenue figures appear in this article. Every result is a percentage, a percentage-point difference, a scale-free metric, a rank, or a count, by design. The evaluation protocol accompanies this preprint as arXiv ancillary files. It contains a written specification of every evaluation choice reported here, and a runnable implementation that takes any panel and any set of forecasters. It also carries the exact scoring code behind the reported numbers, and the experiment scripts as an audit trail. A synthetic panel is included as well, on which the unit-of-analysis result of Section~\ref{sec:inversion} reproduces in direction. Its generator carries the spikiness dial of Section~\ref{sec:dose}, so the appear-and-dissolve result reproduces with no data at all. A reader can therefore run the protocol end to end without our data. The package also carries the scripts behind Section~\ref{sec:m5} --- the M5 replication, the native-hierarchy sweep, and the interval leg --- so those replications are reproducible in full, along with the two failed fragility-statistic designs of Section~\ref{sec:limits}. The 24 benchmarked methods are third-party libraries and one proprietary baseline, and are not re-implemented there.

\textbf{CRediT authorship contribution statement.} \textbf{Md Rezwanul Islam:} Conceptualization, Methodology, Software, Formal analysis, Investigation, Data curation, Writing -- original draft, Writing -- review \& editing, Visualization. \textbf{Wael Mohammed:} Conceptualization, Project administration, Resources, Supervision, Validation, Writing -- review \& editing.

\textbf{Declaration of generative AI and AI-assisted technologies in the writing process.} During the preparation of this work, the authors used Claude (Anthropic) to help organize the structure of the manuscript and to improve the clarity of the writing. After using this tool, the authors reviewed and edited the content and take full responsibility for the final content of the publication.

\end{document}